\documentclass[10pt]{article} 

\usepackage{amsmath,amsfonts,bm}

\def\eqref#1{equation~\ref{#1}}

\def\1{\bm{1}}
\newcommand{\train}{\mathcal{D}}

\def\vx{{\bm{x}}}

\DeclareMathAlphabet{\mathsfit}{\encodingdefault}{\sfdefault}{m}{sl}
\SetMathAlphabet{\mathsfit}{bold}{\encodingdefault}{\sfdefault}{bx}{n}

\def\gG{{\mathcal{G}}}

\def\gO{{\mathcal{O}}}

\newcommand{\R}{\mathbb{R}}

\DeclareMathOperator*{\argmax}{arg\,max}

\usepackage{hyperref}
\usepackage{url}
\usepackage{xcolor}
\usepackage[pdftex]{graphicx}
\usepackage{caption}
\usepackage{subcaption}
\usepackage{cleveref}
\usepackage{booktabs}

\usepackage{algorithm}
\usepackage{algorithmic}
\let\algoAND\AND
\let\AND\classAND
\AtBeginEnvironment{algorithmic}{\let\AND\algoAND}
\usepackage[preprint]{tmlr}

\usepackage{multirow}

\newtheorem{theorem}{Theorem}

\newtheorem{proof}{Proof}

\newcommand\restr[2]{{
  \left.\kern-\nulldelimiterspace 
  #1 
  \vphantom{\big|} 
  \right|_{#2} 
  }}

\title{A Novel Approach to Regularising 1NN classifier for Improved Generalization}

\author{\name Aditya Challa \email adityac@goa.bits-pilani.ac.in \\
      \addr Department of CS\&IS\\
      BITS Pilani KK Birla Goa Campus
      \AND
      \name Sravan Danda \email dandas@goa.bits-pilani.ac.in \\
      \addr Department of CS\&IS \\
      BITS Pilani KK Birla Goa Campus
      \AND
      \name Laurent Najman \email laurent.najman@esiee.fr\\
      \addr Department of Computer Science\\
      Universit\'e Gustave Eiffel
      }

\usepackage[textsize=tiny]{todonotes}

\newcommand{\nseeds}{\texttt{N\_SEEDS}}

\def\month{MM}  
\def\year{YYYY} 
\def\openreview{\url{https://openreview.net/forum?id=XXXX}} 

\begin{document}

\maketitle

\begin{abstract}


In this paper, we propose a class of non-parametric classifiers, that learn arbitrary boundaries and generalize well.

Our approach is based on a novel way to regularize 1NN classifiers using a \emph{greedy} approach. We refer to this class of classifiers as \emph{Watershed Classifiers}. 1NN classifiers are known to trivially over-fit but have very large VC dimension, hence do not generalize well. We show that watershed classifiers can find arbitrary boundaries on any dense enough dataset, and, at the same time, have very small VC dimension;  hence a watershed classifier leads to good generalization. 

Traditional approaches to regularize 1NN classifiers are to consider $K$ nearest neighbours. Neighbourhood component analysis (NCA) proposes a way to learn representations consistent with ($n-1$) nearest neighbour classifier, where $n$ denotes the size of the dataset. In this article, we propose a loss function which can learn representations consistent with watershed classifiers, and show that it outperforms the NCA baseline. 

\end{abstract}

\section{Introduction}
\label{sec:intro}

Deep learning classifiers have obtained state-of-the-art results on the classification problem, but they predominantly use a parametric linear classifier as the classification layer. At the opposite end of the spectrum of classifiers to the linear classifiers lies non-parametric classifiers, such as K-Nearest-Neighbours \cite{Hastie2009}. To our knowledge, non-parametric classifiers and their counterpart loss functions have not been used recently for classification. 

However, non-parametric class of classifiers form an aspect of ML systems with a lot of flexibility to learn arbitrary boundaries. It is known that 1NN classifiers trivially overfit the data. Traditionally, one uses K Nearest Neighbors to regularize this. \cite{DBLP:conf/nips/GoldbergerRHS04} proposes a loss function -- Neighbourhood Component Analysis (NCA), to learn KNN classifiers. \emph{Is there a better way to regularize 1NN classifiers?}

\paragraph{Contributions:} (i) In this article, we propose a novel way to regularize 1NN classifiers using a greedy approach. This class of classifiers are referred to as \emph{Watershed Classifiers}. (ii) Interestingly, the only hyperparameter for this class is \texttt{N\_SEEDS} which directly controls the VC dimension (see \cref{sec:watershedclf}). Moreover, even with a small VC dimension the watershed classifier can learn arbitrarily complex boundaries, assuming that the data is dense enough. This ensures a very small generalization gap theoretically. (iii) To suitably learn the embedding, we also propose a novel loss function. Although the loss function is highly non-convex, SGD approaches worked surprisingly well, as we show in \cref{sec:experiments}. (iv) As delineated in \cref{sec:experiments}, the proposed classifier surpasses NCA in performance. To the best of our knowledge, this represents the first occurrence in which a non-parametric classifier matches or surpasses the accuracy of a linear classifier.

\section{Related Works}

\paragraph{NCA and Metric Learning:} As stated in \cite{DBLP:conf/nips/GoldbergerRHS04}, NCA learns the representations consistent with KNN classifiers. However, the approach does not scale well to large datasets.  In \cite{DBLP:conf/eccv/WuEY18} the authors make suitable modifications to scale this approach to large datasets. Nevertheless, NCA based approaches usually do not lead to state-of-the-art results when compared to linear classifiers. Hence, to our knowledge, NCA based approaches have not been widely used in the context of classification in favour of linear classifiers. However, these have been widely used for \emph{metric learning}. \cite{DBLP:conf/eccv/TehDT20} proposes \texttt{ProxyNCA++} which uses proxy points in the embedding space, based on \cite{DBLP:conf/iccv/Movshovitz-Attias17}. In this article, our focus is mainly on the classification aspect and do not consider the metric learning as an objective.

\paragraph{Relation with Watersheds:} Watersheds have been widely used for image classification and related tasks before the deep learning approaches. To our knowledge, the authors in \cite{falcao1999image,DBLP:journals/pami/CoustyBNC10, DBLP:journals/pami/CoustyBNC09} are the first to propose watersheds on edge-weighted graphs. The authors in \cite{DBLP:journals/spl/ChallaDSN19} propose to use watersheds for classification, but do not learn the representations. In \cite{DBLP:journals/tgrs/ChallaDSN22}, the authors try to learn representations consistent with watershed using Triplet loss as a proxy loss function. As the authors state, their method depends heavily on the graph. In \cite{DBLP:conf/nips/TuragaBHDS09,DBLP:conf/iccv/WolfSKH17} the authors propose novel loss functions to learn watershed segmentation for images. In \cite{DBLP:conf/nips/SanmartinDH19,DBLP:conf/nips/SanmartinDH21} the authors propose (directed) probabilistic watersheds which assumes an underlying KNN graph. All these approaches work for segmenting images, but do not fare well for other tasks. In contrast, we provide a complexity analysis (\cref{sec:watershedclf}), propose a more consistent loss function (\cref{sec:train_nn_watershed}) to learn the representations for multi-class classification. Our approach implicitly learns the graph, and works for other tasks than image segmentation. 

\paragraph{Ultrametric Learning:} Closely related to learning watersheds is the problem of learning ultrametrics. Learning ultrametrics is related to unsupervised clustering of data, and there exists a large body of work \cite{DBLP:journals/jmlr/AckermanB16,DBLP:conf/stoc/Dasgupta16,DBLP:conf/nips/YarkonyF15} exploring the links. In \cite{DBLP:conf/nips/ChierchiaP19} the authors assume an underlying weighted graph and propose a cost function to learn the closest ultrametric. Most of the related approaches assume that the underlying graph is known. In contrast, the loss function proposed here implicitly learns a graph, specifically a minimum spanning tree, for the purposes of classification. 

\section{Watershed Classifier (a.k.a. greedy 1-NN classifier)}
\label{sec:watershedclf}

\begin{figure*}[tb]
\vskip 0.2in
\begin{center}
\begin{subfigure}{0.33\columnwidth}
    \centerline{\includegraphics[width=\columnwidth]{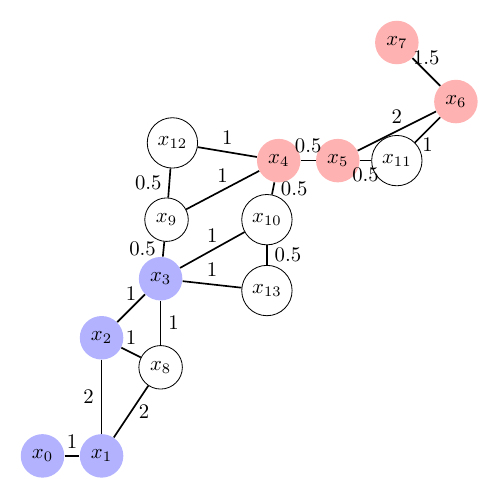}}
    \caption{}
    \label{fig:1a}
\end{subfigure}%
\hfill%
\begin{subfigure}{0.33\columnwidth}
    \centerline{\includegraphics[width=\columnwidth]{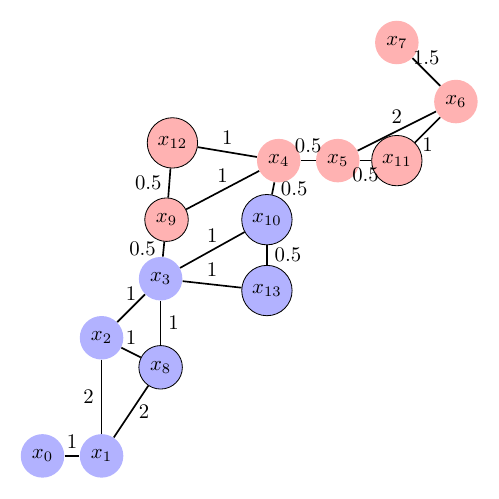}}
    \caption{}
    \label{fig:1b}
\end{subfigure}
\hfill%
\begin{subfigure}{0.33\columnwidth}
    \centerline{\includegraphics[width=\columnwidth]{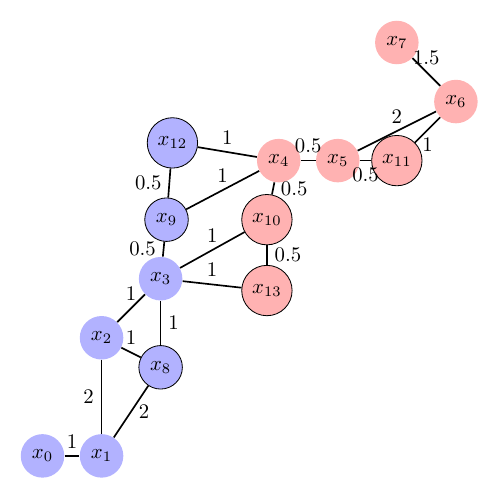}}
    \caption{}
    \label{fig:1c}
\end{subfigure}
\caption{Illustrating the labelling preference using watershed classifier. (a) illustrates an arbitrary set of data points. {\color{blue!20} Blue} dots indicate class 0, {\color{red!20} Red} dots indicate class 1. The unlabelled points have no colour. Few selected edges with corresponding edge-weights are included. (b) and (c) indicates two different labelling. Observe that \textsc{Margin} of (b) is $0.5$, while margin of (c) is $1$. Hence $(c)$ is considered a better labelling than (b) as per the \textsc{Maximum Margin Principle}. }
\label{fig:1}
\end{center}
\vskip -0.2in
\end{figure*}

\begin{figure}[tb]
\vskip 0.2in
\begin{center}
\centerline{\includegraphics[width=0.33\columnwidth]{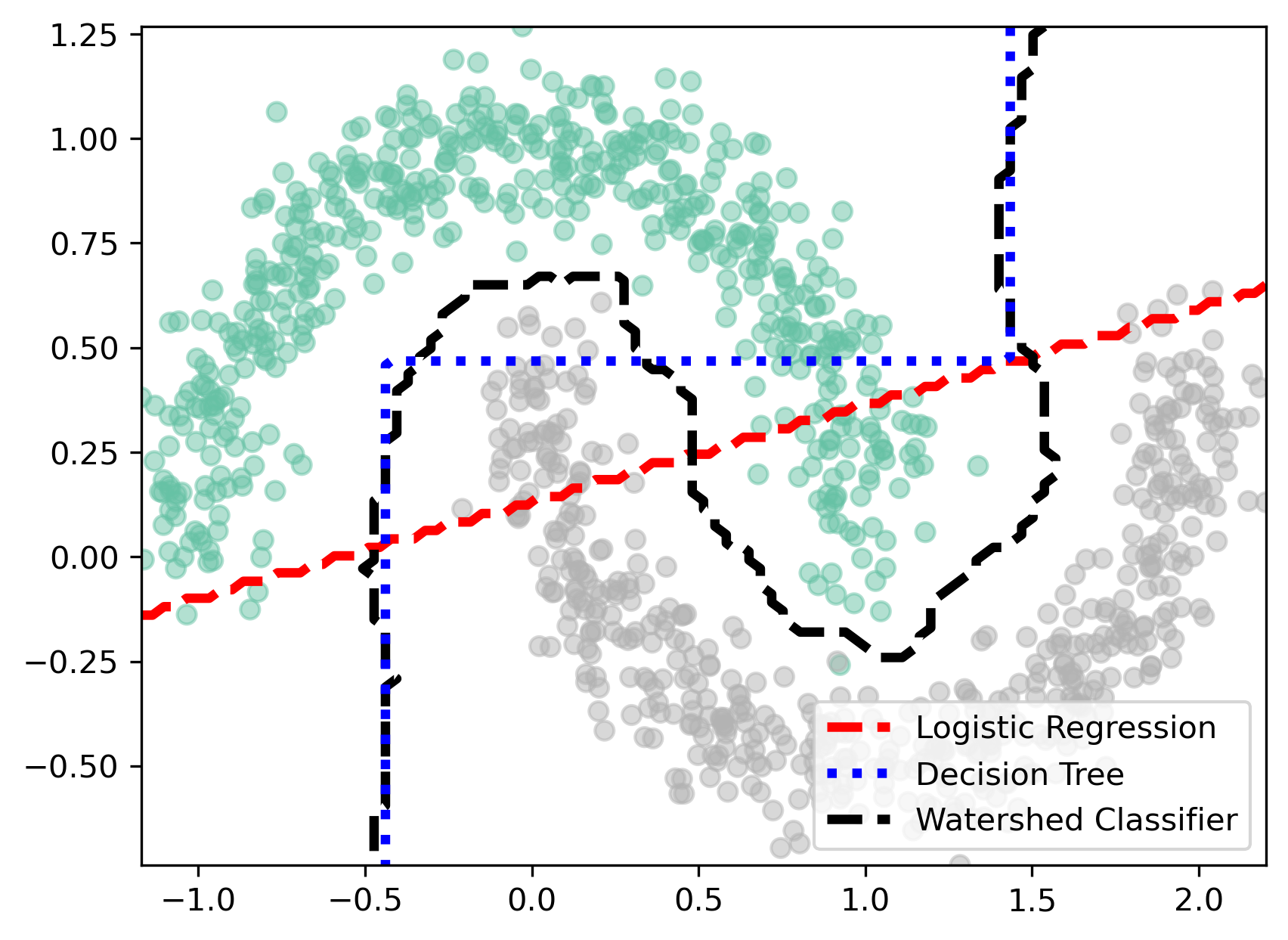}}
\caption{Comparison of Watershed Classifier with Linear Classifier and Decision Tree. Considering a simple toy example, and a fixed VC-dimension of $3$, observe that watershed classifier can find the right boundary.}
\label{fig:2}
\end{center}
\vskip -0.2in
\end{figure}

Let $\train = \{(\vx_i, y(\vx_i)\}$ denote the dataset where $\vx_i \in \R^d$ and $y(\vx_i)$ denote the corresponding labels. We assume that $\train$ consists of both labelled and unlabelled samples. So, we assume that $y(\vx) \in \{0,1,\cdots,K\}$ (classes) for labelled samples and $y(\vx) = -1$ for unlabelled samples. \textbf{Remark:} For developing the theory, we assume $y(\vx) \in \{0,1\}$ (binary classification) for simplicity. In practice, we extend this to multi-class using one-vs-rest approach. 

We implicitly assume a \emph{complete} graph on $\train$ -- $\gG_{\train} = (V, E, W)$, where (i) each vertex corresponds to the data point, $V = \{\vx_i\}$ (ii) $E = V \times V$ and (iii) $W : E \to \R^{+}$ denotes the edge-weights which are restricted to be positive and denote the dissimilarity between the points. Unless otherwise specified, we use $W(\vx_i, \vx_j) = \|\vx_i - \vx_j\|$ as the measure of dissimilarity.

The aim of the classifier is to assign labels to all the unlabelled points -- $\{\vx \mid y(\vx)=-1\}$. Given any arbitrary labelling on the entire dataset -- $\tilde{y}(\vx) \in \{0,1,\cdots K\}$, define \textsc{Margin} as follows:
\begin{equation}
    \textsc{Margin} = \min_{\tilde{y}(\vx_i) \neq \tilde{y}(\vx_j)} W(\vx_i, \vx_j)
\end{equation}
that is, the edge-weight between the ``closest'' pair of points with different labels. 

The \textsc{Maximum Margin Principle} then states -- 
\begin{quote}
    The right labelling for the unlabelled points is obtained from the labelling with the largest \textsc{Margin}.
\end{quote}
 \cref{fig:1} illustrates the preference of the labelling with respect to the \textsc{Maximum Margin Principle}. The above definitions are straight-forward extensions to general edge-weighted graphs of the margin and maximum margin principle defined for support vector machines \cite{Scholkopf2002}.

\paragraph{The above classification rule is the same as greedy-1NN-classification:} There exists a simple rule to obtain the maximum margin labelling -- \emph{From the set of unlabelled points, identify the point which has the smallest edge-weight to the labelled set and label it accordingly. And repeat this step until all points are labelled.} This is the same as performing \emph{greedy-1NN-classification}. \textbf{Remark}: Ties are broken arbitrarily. 

In the example of \cref{fig:1a} -- (i) We first label $\vx_9$ as {\color{blue!50} blue}, (ii) label $\vx_{10}$ as {\color{red!75} red} (iii) We first label $\vx_{12}$ as {\color{blue!50} blue}, (iv) label $\vx_{13}$ as {\color{red!75} red}. This gives the maximum margin labelling.

\paragraph{Notation:} Observe that the data is assumed to contain both labelled and unlabelled samples. To differentiate between these, we refer to the labelled samples as \texttt{SEEDS}. Also, we implicitly assume that there exists \emph{at least one} labelled sample per class in the set of \texttt{SEEDS}. Let \nseeds{} denote the number of labelled samples in each class. \textbf{Remark:} Although it is not necessary that the number of seeds be the same for all the classes, we assume so for the sake of simplicity.

\begin{table*}[tb]
\caption{Comparison of Watershed Classifiers with Linear Classifier and Decision Trees}
\label{table:1}
\vskip 0.15in
\begin{center}
\begin{small}
\begin{tabular}{p{0.24\columnwidth}p{0.24\columnwidth}p{0.24\columnwidth}p{0.2\columnwidth}}
\toprule
{\sc Property} & {\sc Watershed Classifier} & {\sc Linear Classifier} & {\sc Decision Trees} \\
\midrule
{\sc Complexity} & \nseeds{} & Number of features & Number of Splits \\
{\sc Non-Linear Boundaries} & Yes & No & Axis-Aligned-Steps \\
{\sc Computation} & $\gO(n \log(n))$ & $\gO(n)$ & $\gO(n)$ \\
{\sc Training} & Building a space partitioning data structure. & Fitting the parameters of the hyperplane. & Identify the best splits for each feature. \\
{\sc Inference for 1 sample} & 1-Nearest-Neighbor & Classification rule & Parsing the decision tree  \\
{\sc Inference for $n$ samples} & Minimum Spanning Tree & Classification rule & Parsing the decision tree  \\
{\sc Inference Depends on other samples} & Yes & No & No \\
\bottomrule
\end{tabular}
\end{small}
\end{center}
\vskip -0.1in
\end{table*}

\subsection{Complexity Analysis of Watershed Classifier:}

Note that watershed classifier has \emph{no trainable parameters} and has only one hyperparameter \nseeds{}. In this section, we analyse the complexity of the watershed classifier.

Recall that VC-Dimension is defined as the size of the largest set of which the algorithm can shatter \cite{books/daglib/0033642}. We then have the following result -- 
\begin{theorem}
 Let $\{\vx_i\}$ denotes the set of data points and $\gG_{\train} = (V,E,W)$ denotes the complete graph. Assume that $W(\vx_i, \vx_j) \neq W(\vx_k, \vx_l)$ for all $i\neq k$ or $j\neq l$. That is, all edge weights are assumed to be distinct. Let the number of classes be $2$ (binary classification). Then, VC-dimension of the watershed classifier with \nseeds{} is $2\times \nseeds{}$.
\end{theorem}
\begin{proof}
    The proof follows from the fact that -- (a) if all edge-weights are distinct, then the 1-Nearest Neighbour is unique for all points and (b) Once the seeds are fixed, then the labelling is unique. 

    Let $k$ denote the number of points such that all possible configurations in $\{0,1\}^k$ can be obtained using the watershed classifier. Observe that $k \leq 2 \times \nseeds{}$ since, when the $2 \times \nseeds{}$ are fixed, the labels of the other data points are fixed and unique. And it is easy to see that all possible configurations can be obtained when $k=2\times \nseeds{}$. Hence, the VC dimension is $2 \times \nseeds{}$.
\end{proof}

Although the above result is relatively simple, there are a few significant implications - (i) The only hyper-parameter $\texttt{N\_SEEDS}$ directly control the VC dimension, and hence generalization gap. (ii) If the features are dense enough, then by simply taking $\texttt{N\_SEEDS}=1$ can give very good generalization bounds. Apart from this the watershed classifier also exhibits several properties which are different from any of the existing classifiers.

\paragraph{Comparison with existing classifiers:} \cref{table:1} shows the summary of the comparison of watershed classifiers with linear classifiers and decision trees. Several properties of the watershed classifier are inherited from the nearest-neighbour classifier. Training involves building a space partitioning data structure for efficient similarity search \cite{DBLP:conf/www/DongCL11,DBLP:journals/tbd/JohnsonDJ21}. Inference for a single sample is based on a simple 1-nearest-neighbor search. 

Watershed classifier differs from nearest-neighbours for inference on several samples. Note that Linear Classifiers, Decision Trees and KNN classifiers do not distinguish between a single sample vs multiple sample inference. In case of watershed classifier, one should classify the \emph{closest points} before classifying other points. This is similar to the active learning framework \cite{DBLP:journals/corr/abs-2112-07963} except that 1NN rule provides an efficient way and removes the high computational training of the models. 

Regarding complexity, it is known that nearest neighbour approaches are prone to overfitting, with 1NN classifier which overfits in a single shot. Traditional approaches use KNN to reduce the complexity. Watershed classifier, unconventionally, uses a greedy approach to reduce the overfitting. This approach to reducing complexity is probably the key reason why watershed out-performs linear classifiers when working with neural networks (See \cref{sec:experiments} for details). 

Linear classifiers have a complexity which scales as $\gO(d)$, where $d$ is the number of features. However, the complexity of Watershed classifier does not depend on the embedding dimension. Decision stumps (Decision Trees with a single split) only search for axis aligned splits \footnote{While it is possible to search for non-axis aligned splits \cite{DBLP:conf/ijcai/HeathKS93}, most of the current state-of-art approaches rely on axis-aligned splits}. Interestingly, the VC-dimension of a decision stump is $2$, which is equal to the VC-dimension of the Watershed classifier with $\nseeds{}=1$. However, Watershed classifier is capable of having non-linear boundaries.

Regarding computational requirement, watershed classifier requires\footnote{Greedy 1NN search is similar in complexity to constructing a Euclidean minimum spanning tree. \cite{DBLP:conf/soda/AryaM16} shows that $\epsilon-$approximate EMST for $n$ points can be found in $\gO(n\log(n) + (\epsilon^{-2}\log^2(1/\epsilon))n$ time.} $\gO(n\log(n))$ to classify $n$ points. However, both linear classifier and Decision Trees have a lower complexity ($\gO(n)$) comparatively. 

A property which makes watershed classifier unique compared to the rest, is the fact that the labelling actually depends on \emph{density of the samples}. In presence of very high noise in the features, Watershed classifier does not work well. In comparison, both linear models and decision trees are capable of handling noisy inputs comparatively well. However, this issue is easily rectified when using watershed classifier along with neural nets (see \cref{sec:train_nn_watershed}).

\cref{fig:2} illustrates the difference between Watershed classifier with Linear and Decision Tree Classifiers. Even with the small VC-dimension of $3$, we have that watershed classifier can identify the non-linear boundary. (See \cref{sec:append1} for more details about the visualization of the boundaries).

\section{Learning Neural Networks to get Watershed Representations}
\label{sec:train_nn_watershed}
\begin{algorithm}[tb]
   \caption{Computing Watershed Loss}
   \label{alg:watershedloss}
\begin{algorithmic}[1]
   \STATE {\bfseries Input:} Representations $\textsc{X\_rep}=\{f_{\theta}(\vx_i)\}$, Labels $\{y_i\}$, Hyper parameter \nseeds{}
   \STATE For each class, select \nseeds{} points randomly from $\textsc{X\_rep}$. We refer to this set as \textsc{X\_seeds}.
   \STATE Propagate the labels from the \textsc{X\_seeds} to the rest of data points. 
   \STATE Let \textsc{X\_correct} denote the set of points which have the correct labels. (\textbf{Remark:} Note that the propagated labels may/may-not match with the input labels $y_i$.  Clearly, $\textsc{X\_seeds} \subseteq \textsc{X\_correct}$.)
   \STATE Further, let \textsc{X\_correct\_l} denote the subset of correctly labelled point with label \textsc{l}. \textbf{Remark:} Thus we have,
   \begin{equation}
       \textsc{X\_correct} = \bigcup_{\textsc{l}} \textsc{X\_correct\_l}
   \end{equation}
   \STATE Let \textsc{1NN}($\vx_i$, \textsc{X\_correct\_l}) denote the 1 nearest neighbor of the point $x_i$ in  \textsc{X\_correct\_l}. For each data point $\vx_i$ compute the probabilities and the corresponding loss as -
   \begin{equation}
   \small 
   \begin{aligned}
       & \vx_{i,1nn,L} =  \textsc{1NN}(f_{\theta}(\vx_i), \textsc{X\_correct\_l} \setminus \{f_{\theta}(\vx_i)\}) \\
       &p_{i,L} \propto \exp \left(-|| f_{\theta}(\vx_i) - f_{\theta}(\vx_{i,1nn,L}) ||\right) \\
       &\textsc{Loss}(\vx_i) = \sum_{k=1}^{K} I[y_i=K] \log(p_{i,K})
   \end{aligned}
   \end{equation}
   where $I[.]$ denotes an indicator function.
   \STATE \textbf{return} sum of losses for all data points.
\end{algorithmic}
\end{algorithm}

\begin{figure*}[tb]
\vskip 0.2in
\begin{center}
\centerline{\includegraphics[width=\columnwidth]{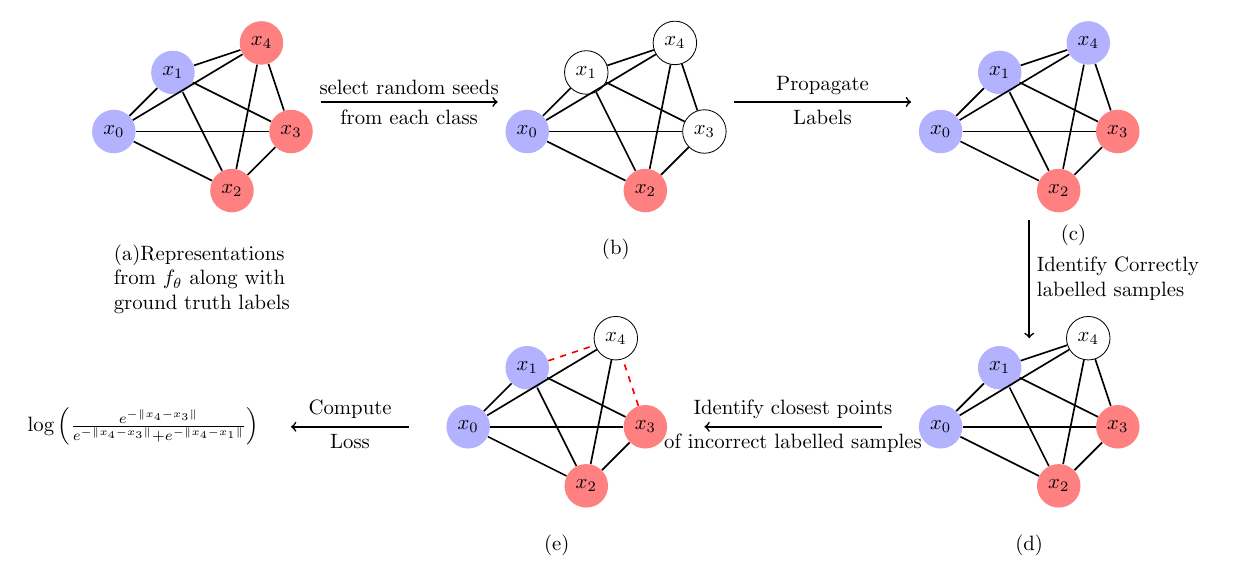}}
\caption{Illustrating the computation of watershed loss. (a) shows the toy representations obtained from $f_{\theta}$ along with the ground-truth labels. (b) shows the seeds selected from each class. We assume $\nseeds{}=1$ in this case. (c) illustrates the propagation of the labels. Note that $x_4$ has a ground-truth label of {\color{red!75}red}, but is labelled {\color{blue!50}blue} with label propagation. All the other labels match the ground-truth. (d) identifies the correctly labelled samples. (e) identifies closest samples which are correctly labelled. In this case, for $x_4$, the closest  {\color{blue!50}blue} sample is $x_1$ and the closest {\color{red!75}red} sample is $x_3$. Note that, one should compute this for all samples. We only choose $x_4$ here for illustration. Finally, we compute the loss for each sample using \cref{eq:loss1} and \cref{eq:loss2}.}
\label{fig:3}
\end{center}
\vskip -0.2in
\end{figure*}

Recall that the Watershed Classifier in \cref{sec:watershedclf} does not have any trainable parameters and hence is not powerful enough to handle highly noisy datasets. To rectify this issue, in this section, we provide a way to combine the Watershed classifier with neural nets. Formally, we  wish to solve
\begin{equation}
    \argmax_{\theta} \text{Acc}(\text{Watershed Classifier}(f_{\theta}(X))
\end{equation}
Here $f_{\theta}$ denotes the neural network architecture which learns the representations, $X = \{\vx_i\}$ denotes the input data-points. In words -- we wish to learn the representations of the data such that when watershed is applied we get as large accuracy as possible.

\paragraph{Watershed Loss Function:} We propose a corresponding loss function described in \cref{alg:watershedloss} to train the network $f_{\theta}$. Computing the loss function consists of three stages -- (i) Propagating the labels (ii) Selecting the ``closest'' correctly labelled sample from each class (iii) Computing the cross-entropy loss with respect to the closest labels.

\paragraph{(i) Propagating watershed labels:} Let $X=\{\vx_i\}$ denote the input data-points, and $f_{\theta}(X)$ denote the representations of the data-points. First select a \emph{random} subset of points of size \nseeds{} from each class. Using this subset of points as seeds, propagate the labels using greedy 1-Nearest-Neighbor approach. 

\paragraph{(ii) Selecting the closest correctly labelled sample from each class:} Note that the propagated labels may or may not match the true labels. Consider only the subset of points which are correctly labelled -- \textsc{X\_correct}. Also consider the subset of points which are correctly labelled and has a class \textsc{l} -- \textsc{X\_correct\_l}. For each data point, we select the closest point as follows:
\begin{equation}
    \vx_{i,1nn,L} =  \textsc{1NN}(f_{\theta}(\vx_i), \textsc{X\_correct\_l} \setminus \{f_{\theta}(\vx_i)\})
\end{equation}
Here \textsc{1NN} denote the computation of the 1-nearest-neighbor in the set \textsc{X\_correct\_l} excluding itself ($f_{\theta}(\vx_i)$).

\paragraph{(iii) Computing the loss:} Finally, the loss is computed akin to the cross entropy loss. We construct the probabilities from the Euclidean distances as 
\begin{equation}
    p_{i,L} = \frac{\exp \left(-|| f_{\theta}(\vx_i) - f_{\theta}(\vx_{i,1nn,L}) ||\right)}{\sum_{k=1}^{K}\exp \left(-|| f_{\theta}(\vx_i) - f_{\theta}(\vx_{i,1nn,k}) ||\right)}
    \label{eq:loss1}
\end{equation}
and then construct the cross-entropy loss based on these probabilities:
\begin{equation}
    \text{Loss}(\vx_i) = \sum_{k=1}^{K} I[y_i=k] \log(p_{i,k})
    \label{eq:loss2}
\end{equation}
where $K$ denotes the number of classes.

A simple illustration of the computation of the loss function is given in \cref{fig:3}.

\paragraph{Intuition behind the loss function:}

Why does the above loss function learn representations consistent with watershed classifier? -- Intuitively, at each stage of optimizing the loss function we -- (i) make sure that the correct propagation is preserved and (ii) wrong propagation of labels is corrected. This ensures that the labels are correctly propagated. So, the correct propagation from the seeds can be considered an invariant which only improves with learning. The experiment in \cref{sec:verify_loss} verifies this aspect of the loss function.

\subsection{Training $f_{\theta}$ using watershed loss}

\paragraph{Watershed loss is Non-Convex:} A generic rule of thumb when designing a loss function for neural networks is -- replacing the neural-net with a simple matrix should result in a convex loss function. This holds true for almost all known widely used loss functions. However, the watershed loss function which is proposed here does not have this property. Interestingly, we empirically find that stochastic gradient descent (SGD) performs surprisingly well on this loss function. 

\paragraph{Does not distribute over samples:} Almost all the loss functions -- $\ell(.,.)$ distribute over the samples, i.e., they can be written as -- $\sum_{i} \ell(f_{\theta}(\vx_i),y_i)$. This does not hold true for watershed loss. Hence, the batch size $B$ plays a crucial role in training the network $f_{\theta}$. In practice -- we first draw $N$ random samples from the dataset, each with size $B$ -- $\{X_{b_i} = \{\vx_{b_i,j}\}_{j=1}^{B},y_{b_i} = \{y_{b_i,j})\}_{j=1}^{B}\}$ and train $f_{\theta}$ using the loss $ L(\theta) =(1/N)\sum_{i=1}^{N} \text{Watershed Loss} (f_{\theta}(X_{b_i}), y_{b_i})$ where the Watershed Loss is obtained using \cref{alg:watershedloss} and $f_{\theta}(X_{b_i}) = \{f_{\theta}(\vx) \mid \vx \in X_{b_i}\}$. 

\paragraph{Evaluating $f_{\theta}$ trained using watershed loss}

To be consistent with the training procedure, we evaluate the network trained by -- (i) First select $N$ random samples of size $B$ each -- $\{X_{b_i} = \{\vx_{b_i,j}\}_{j=1}^{B},y_{b_i} = \{y_{b_i,j})\}_{j=1}^{B}\}$ (ii) For each test sample $\vx_{i}^{+}$ we perform 1NN classifier on each of the sample $B$ and predict the label of $\vx_{i}^{+}$. The final label is taken to be the \emph{most frequent label} in the predictions for each of the samples. 

\begin{table*}[t]
\caption{Watershed Classifier vs NCA vs Linear Classifier on \textsc{FashionMNIST}. $f_{\theta}$ is taken to be the linear embedding with $\texttt{EMBED\_DIM}=4$ or $16$.}
\label{table:2}
\vskip 0.15in
\begin{center}
\begin{small}
\begin{sc}
\begin{tabular}{p{5em}p{5em}cccccccp{5em}}
\toprule
   & {} & \multicolumn{6}{c}{Watershed Classifier} & NCA & Linear Classifier\\
   & N SEEDS &    1   &    5   &    10  &    20  &    40  &    100 & & \\
\midrule
EMBED- DIM & BATCH- SIZE &        &        &        &        &        &        & & \\
\midrule
4  & 1020 & 0.7802 & 0.8211 & 0.8272 & 0.8296 & 0.8320 &    -- & \multirow{3}{*}{0.4375}&\multirow{3}{*}{0.8120} \\
   & 2040 & 0.7960 & 0.8265 & 0.8288 & 0.8300 & 0.8294 & 0.8307 & & \\
   & 4090 & 0.8078 & 0.8252 & 0.8256 & 0.8304 & 0.8296 & 0.8305 & & \\
\midrule
16 & 1020 & 0.8509 & 0.8716 & 0.8739 & 0.8769 & 0.8792 &    -- & \multirow{3}{*}{0.6969}&\multirow{3}{*}{0.8600} \\
   & 2040 & 0.8533 & 0.8743 & 0.8768 & 0.8792 & 0.8838 & 0.8809 & & \\
   & 4090 & 0.8474 & 0.8736 & 0.8740 & 0.8807 & 0.8812 & 0.8815 & & \\
\bottomrule
\end{tabular}
\end{sc}
\end{small}
\end{center}
\vskip -0.1in
\end{table*}

\begin{table*}[t]
\caption{Watershed Classifier vs NCA vs Linear Classifier on \textsc{CIFAR10},\textsc{CIFAR100}, \textsc{FashionMNIST}. $f_{\theta}$ is taken to be three different types of architectures -- \textsc{M3LC}, \textsc{MR18}, \textsc{M3FF}. Observe that while on \textsc{FashionMNIST} all approaches give similar results, Watershed classifier outperforms NCA and works comparably to Linear classifier on \textsc{CIFAR10/100} datasets.}
\label{table:3}
\vskip 0.15in
\begin{center}
\begin{small}
\begin{sc}
\begin{tabular}{llcccc}
\toprule
DATASET & MODEL & Width &Watershed Classifier & NCA & Linear Classifier \\
\midrule
\textsc{FashionMNIST} & \textsc{M3LC} &256  & 0.9229 & 0.9164 & 0.9218 \\
    & \textsc{M3LC} & 64 & 0.9120 & 0.9030 & 0.9089 \\
    \cmidrule{2-6}
   & \textsc{M3FF} & 256 & 0.8918 & 0.8725 & 0.8799 \\
   & \textsc{M3FF} & 64 & 0.8801 & 0.8737 & 0.8760 \\
   \cmidrule{2-6}
   & \textsc{MR18} & & 0.9356 & 0.9217 & 0.9246 \\
\midrule
\textsc{CIFAR10} & \textsc{M3LC} &256  & 0.8616 & 0.8040 & 0.824\\
    & \textsc{M3LC} & 64 & 0.8287 & 0.7359 & 0.789 \\
    \cmidrule{2-6}
   & \textsc{M3FF} & 256 & 0.6268 & 0.5585 & 0.6128 \\
   & \textsc{M3FF} & 64 & 0.5491 & 0.5012 & 0.5439 \\
   \cmidrule{2-6}
   & \textsc{MR18} & & 0.9191 & 0.8512 & 0.870 \\
\midrule
\textsc{CIFAR100} & \textsc{M3LC} &256  & 0.5788 & 0.3199 & 0.5478 \\
    & \textsc{M3LC} & 64 & 0.4899 & 0.2602 & 0.4702 \\
    \cmidrule{2-6}
   & \textsc{M3FF} & 256 & 0.3208 & 0.1841 & 0.3290 \\
   & \textsc{M3FF} & 64 & 0.2516 & 0.1603 & 0.2517 \\
   \cmidrule{2-6}
   & \textsc{MR18} & & 0.6989 & 0.3494 & 0.6066 \\
\bottomrule
\end{tabular}
\end{sc}
\end{small}
\end{center}
\vskip -0.1in
\end{table*}

\section{Experiments and Analysis}
\label{sec:experiments}

The watershed classifier is unique in comparison to other classifiers in two major ways -- (i) Its representational capacity is close to 1NN, i.e., it can learn arbitrarily complex boundaries and (ii) The generalization is however much better than 1NN, and is in-fact comparable to the linear classifiers. In this section, we perform simple experiments to verify these properties. \textbf{Remark:} The code to generate these results is provided in the supplementary material.

\subsection{Verifying Representational Capacity of Watershed Classifier}

\begin{figure}[tb]
\vskip 0.2in
\begin{center}
\centerline{\includegraphics[width=0.9\columnwidth]{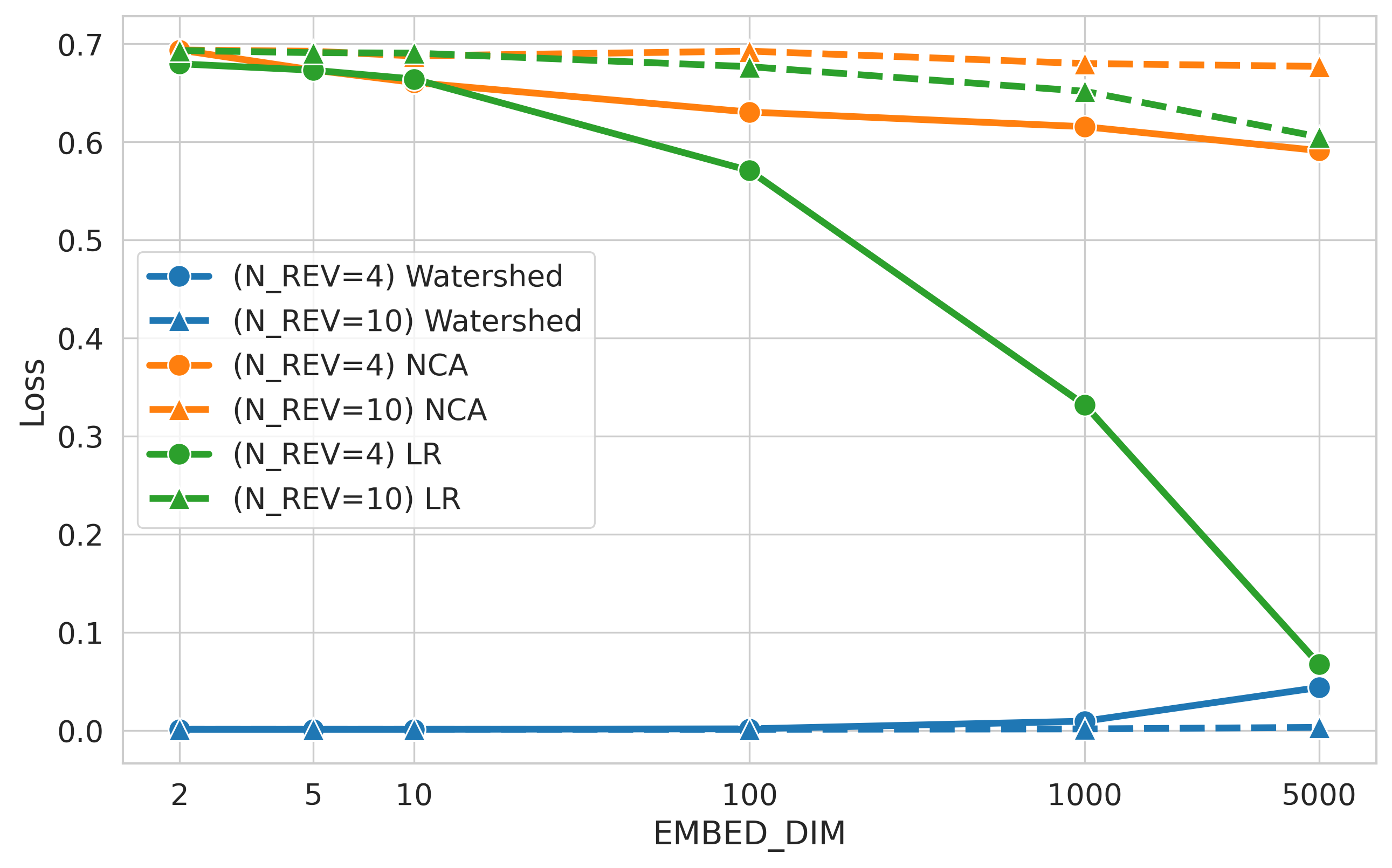}}
\caption{Representation Capacity of Watershed Classifier.}
\label{fig:6}
\end{center}
\vskip -0.2in
\end{figure}

\paragraph{Experiment Design:} Consider the popular \textsc{Spiral Dataset} where we have a parameter \texttt{N\_REV} which dictates how many times the spiral revolves around the origin. Hence, larger the \texttt{N\_REV}, larger the complexity of the boundary. Assume that the embedding network $f_{\theta}$ is a matrix of size $2 \times \texttt{EMBED\_DIM}$. We then investigate -- What is the minimum width $\texttt{EMBED\_DIM}$ required to get 100\% training accuracy? We expect that if the classifier has large representational capacity, then $\texttt{EMBED\_DIM}$ would be small. Otherwise, it would be large. As baselines, we consider the NCA loss \cite{DBLP:conf/nips/GoldbergerRHS04} and the standard linear classifier with the same $f_{\theta}$.

We measure the best loss one can achieve with different values of $\texttt{EMBED\_DIM}$. The loss for watershed is the one computed in \cref{alg:watershedloss}, loss for NCA is computed using the loss proposed in \cite{DBLP:conf/nips/GoldbergerRHS04}, and the loss for linear classifiers is computed using the \emph{binary cross-entropy} loss.  We consider two values of $\texttt{N\_REV}$ -- $4$ and $10$. 

\paragraph{Observations:} The results are plotted in \cref{fig:6}. Illustrations can be found in \cref{sec:spiral_illust}. Firstly, as it is intuitive, with large enough $\texttt{EMBED\_DIM}$, one can get the loss to arbitrary small. However, for $\texttt{N\_REV}=4$ NCA and Linear classifiers require a high width. But watershed classifier can obtain negligible training loss even with $\texttt{EMBED\_DIM}=2$. For $\texttt{N\_REV}=10$, we could not reach $0$ loss, even with $\texttt{EMBED\_DIM}=5000$ for both NCA and Linear classifiers. But, watershed is able to fit the data with $\texttt{EMBED\_DIM}=2$. 

This shows that there is a very high representational capacity for watershed classifier. Indeed, it is known that 1NN classifiers can trivially overfit the dataset, and watershed classifier inherits this property from 1NN classifier. However, the only caveat is that -- We assume that the density of the points is high, as is the case with the \textsc{Spiral dataset} considered here. 

\subsection{Generalization for Linear Embedding}

\paragraph{Experiment Design:} To investigate the generalization performance of the watershed classifier, we consider the case where $f_{\theta}$ is a linear embedding, and use \textsc{FashionMNIST} dataset. We consider two values of $\texttt{EMBED\_DIM}$ -- $4$ and $16$. As baselines, we compare the performance with NCA and linear classifiers. \cref{table:2} shows the results obtained.

\paragraph{Observation 1 -- Dependence on \texttt{N\_SEEDS}:} Recall that \texttt{N\_SEEDS} controls the complexity of the watershed classifier. Hence, for very small \texttt{EMBED\_DIM}, having a high complex classifier is expected to increase the accuracy. This is observed in \cref{table:2} -- where, with $\texttt{EMBED\_DIM}=4$, we obtain a score of $\approx 0.79$ with $\texttt{N\_SEEDS}=1$, and increases to $\approx 0.83$ with $\texttt{N\_SEEDS}=100$. For comparison, the linear classifier obtains a score of $0.81$ with the same \texttt{EMBED\_DIM}.

\paragraph{Observation 2 -- Dependence on \texttt{BATCH\_SIZE}:}  Watershed classifier expects that the representations are dense to work effectively. When training large values of \texttt{BATCH\_SIZE}, this can be achieved efficiently, and hence we see a slight increase in the score when considering $\texttt{EMBED\_DIM}=4$ and increasing the \texttt{BATCH\_SIZE} from $1020$ to $4090$. However, for a higher $\texttt{EMBED\_DIM}=16$, there is not much effect, since dense manifolds can be learned much more effectively with larger \texttt{EMBED\_DIM}.

\paragraph{Observation 3 -- Comparison with NCA:} Both NCA and Watershed classifier learns representations consistent with KNN classifiers. However, the strategy for regularization is different -- NCA considers $K$ nearest neighbours while Watershed considers a greedy approach. \cref{table:2} shows that greedy regularization is clearly superior to the alternate of considering $K$ nearest neighbours.

Tuning \texttt{N\_SEEDS}, we achieve the best score of $0.83$ with $\texttt{EMBED\_DIM}=4$, and $0.883$ with $\texttt{EMBED\_DIM}=16$. In comparison, NCA achieves $0.44$ and $0.69$, respectively. This illustrates two facts about watershed classifier -- (i) Since the VC dimension is low, the generalization is much better and (ii) While both NCA and Watershed Classifier requires the representations to be dense, Watershed classifier can also work effectively with lesser dense representations. To see this, consider the difference between the scores -- we get a difference of $0.4$ for $\texttt{EMBED\_DIM}=4$ and $0.18$ for $\texttt{EMBED\_DIM}=16$. 

\paragraph{Observation 4 -- Comparison with Linear Classifiers:} Linear Classifiers and Nearest Neighbour classifiers can be considered as two opposing strategies for classification \cite{Hastie2009}. Since Watershed classifier is inherited from Nearest Neighbour classifier, these classes share a lot of properties. On the other hand, while Linear classifiers try to separate the classes with linear boundaries, Watershed classifiers try to obtain dense representations so that the margin between the classes is maximized. Interestingly, although Watershed classifier outperforms the Linear classifier by a few points, we observe that both Linear classifiers and Watershed classifier perform similarly well. 

\subsection{Performance with large embedding networks $f_{\theta}$}
\label{ssec:large_network}

Finally, we validate the watershed classifier with a large embedding networks $f_{\theta}$ in this section. More details about the experiments in this section can be found in \cref{sec:exp_largenetwork}.

\paragraph{Experiment Design:} To maintain diversity we consider three different architectures -- (i) \textsc{M3LC} -- A 3 layer convolution network, (ii) \textsc{M3LFF} -- A 3 layer fully connected feed-forward network, (iii) \textsc{M3LR18} -- Resnet18 architecture modified for \textsc{CIFAR} datasets. Further, we also use widths $64$ and $256$ for \textsc{M3LC} and \textsc{M3LFF} networks. We use 3 datasets -- \textsc{FashionMNIST}, \textsc{CIFAR10} and \textsc{CIFAR100} for comparison. Results can be found in \cref{table:3}.

\paragraph{Observations:} Most of the observations are consistent with the linear embedding networks -- Watershed works on par with the linear classifiers, but out-performs NCA by a large extent. Thus, providing further evidence that greedy regularization of 1NN is more effective than KNN approaches.  

\paragraph{Conjecture on complexity of Linear Classifier:}  It is surprising that linear classifiers work on par with greedy regularization of 1NN classifiers, although coming from two opposite ends of the spectrum \cite{Hastie2009}. While the linear classifier has VC dimension which scales with the \texttt{EMBED\_DIM}, the watershed classifier has a VC dimension of \texttt{N\_SEEDS}. Specifically, we observe that considering \texttt{N\_SEEDS} to be between $1$ and $5$ watersheds perform similarly well to linear classifiers. Thus, we conjecture that -- \emph{The complexity of Linear classifiers actually scale with the number of classes and not with \texttt{EMBED\_DIM}}. The reason we expect is that -- When learning a linear classifier, we are actually searching for a subspace with dimension -- \texttt{N\_CLASSES}. However, the VC dimension does not consider the optimization aspect and hence over-estimates the complexity of the classifier.

\section{Conclusion and Future work}


Parametric linear classifiers have largely been used with deep networks to obtain state-of-the art results. In comparison, to our knowledge, Non-parametric methods such as KNN do not obtain results comparable to linear classifiers. Neighbourhood Component Analysis (NCA) aims to learn the embedding consistent with KNN classifier. However, as we argue in this article -- \emph{It is better to regularize 1NN classifier using greedy approach rather than K-Nearest Neighbour approach.}

Using the greedy regularization, we propose \emph{Watershed Classifiers}. Watershed classifiers have a single hyperparameter -- \texttt{N\_SEEDS}, which control the VC dimension. We illustrate that, even with small VC-dimension, we can fit arbitrarily complex boundaries, assuming the high enough density. We propose a loss function in \cref{alg:watershedloss} to train the embedding. Interestingly, we observe that SGD approaches work very well, even if the loss function itself is highly non-convex. In section \cref{sec:experiments}, we show that watershed classifiers outperform the NCA for classification, and either outperform or match the performance of linear classifiers. To our knowledge, this is the first time a non-parametric classifier could match or outperform the parametric linear classifier.





\section{Broader impact}

This paper presents work whose goal is to advance the field of Machine Learning. There are many potential societal consequences of our work, none of which we feel must be specifically highlighted here.





\bibliography{references}
\bibliographystyle{tmlr}

\newpage
\appendix
\onecolumn

\section{Verifying the Watershed Loss Function}
\label{sec:verify_loss}

\begin{figure*}[ht]
\vskip 0.2in
\begin{center}
\begin{subfigure}{0.33\columnwidth}
    \centerline{\includegraphics[width=\columnwidth]{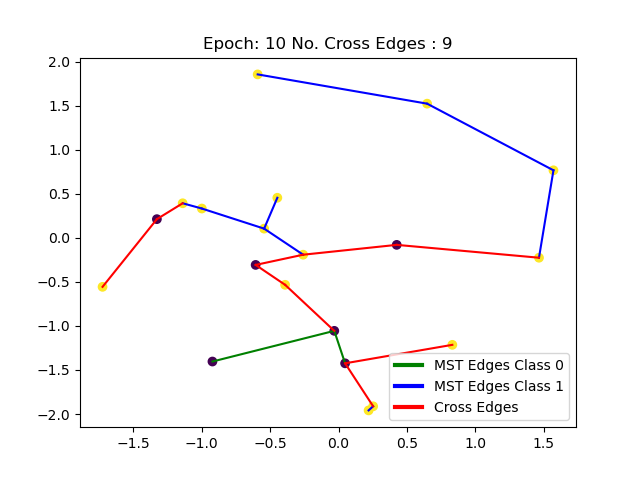}}
    \caption{}
    \label{fig:4a}
\end{subfigure}%
\hfill%
\begin{subfigure}{0.33\columnwidth}
    \centerline{\includegraphics[width=\columnwidth]{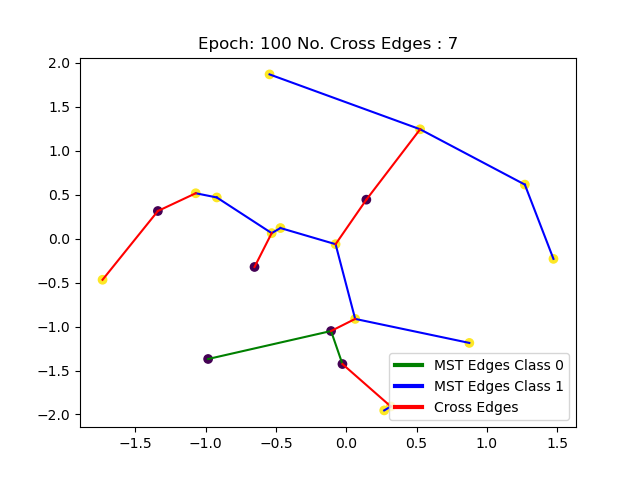}}
    \caption{}
    \label{fig:4b}
\end{subfigure}
\hfill%
\begin{subfigure}{0.33\columnwidth}
    \centerline{\includegraphics[width=\columnwidth]{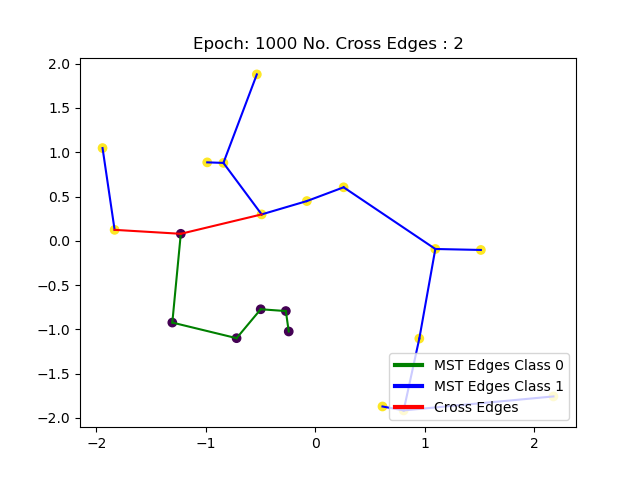}}
    \caption{}
    \label{fig:4c}
\end{subfigure}
\caption{Illustrating that the loss function in \cref{sec:train_nn_watershed} indeed is consistent with greedy 1NN propagation. The purple dots indicate class $1$ and the yellow dots indicate class $0$. Observe that as training progresses we have that number of cross-edges in the minimum spanning tree reduces. Note that greedy 1NN propagation is similar to the Prim's algorithm for constructing a minimum spanning tree. And hence the number of cross-edges provide a good measure on the efficacy of propagation.}
\label{fig:4}
\end{center}
\vskip -0.2in
\end{figure*}

Here, we empirically verify that the loss function in \cref{sec:train_nn_watershed} improves the correct label propagation. 

\paragraph{Experimental Setup:} Consider $20$ arbitrary points on a 2d plane consisting of both class $0$ and class $1$. To experimentally verify the loss function, we directly optimize the co-ordinates of the points. We consider the simple SGD optimizer with learning rate $0.1$ and no momentum.

\paragraph{Measuring the efficacy of propagation:} To measure if the propagation is happening correctly, we look at number of cross edges -- edges for which the end-points do not have the same labels, on the minimum spanning tree. Note that Greedy-1NN propagation is similar to the Prim's algorithm for constructing the minimum spanning tree. And hence, the number of cross-edges capture the efficacy of propagation quite well.

\cref{fig:4} shows the result of optimizing the loss function. Observe that at epoch $10$ we have $9$ cross edges, at epoch $100$ we have $7$ and at epoch $1000$ we only have 2 cross edges. This verifies that the loss function indeed learns the representations which are consistent with greedy 1NN propagation. 

\section{Remark on \cref{fig:2}} 
\label{sec:append1}

We consider the two moons dataset\footnote{\url{https://scikit-learn.org/stable/modules/generated/sklearn.datasets.make_moons.html}} from scikit-learn with $1000$ samples, $\texttt{noise}=0.1$ and $\texttt{random\_state}=0$. Both the linear classifier and the decision tree classifier are also implemented using scikit-learn package. 

Recall that inference for multiple samples differs in case of watershed classifiers and other classifiers. For visualization of the boundary, one should estimate the class for each point in the entire 2d grid. It is imperative that all the grid points are not considered together in the case of watershed classifier. This is because -- A random sample technically should be from the same distribution as the training data. Considering all the grid-points would violate this assumption.


\section{\textsc{Spiral Dataset} Illustrations}
\label{sec:spiral_illust}

\begin{figure*}[tb]
\vskip 0.2in
\begin{center}
\begin{subfigure}{0.35\columnwidth}
    \centerline{\includegraphics[width=\columnwidth]{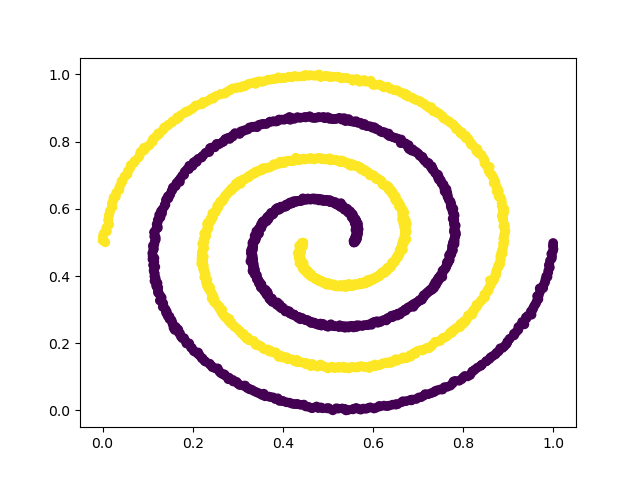}}
    \caption{Spiral dataset}
    \label{fig:5a}
\end{subfigure}%
\hfill%
\begin{subfigure}{0.35\columnwidth}
    \centerline{\includegraphics[width=\columnwidth]{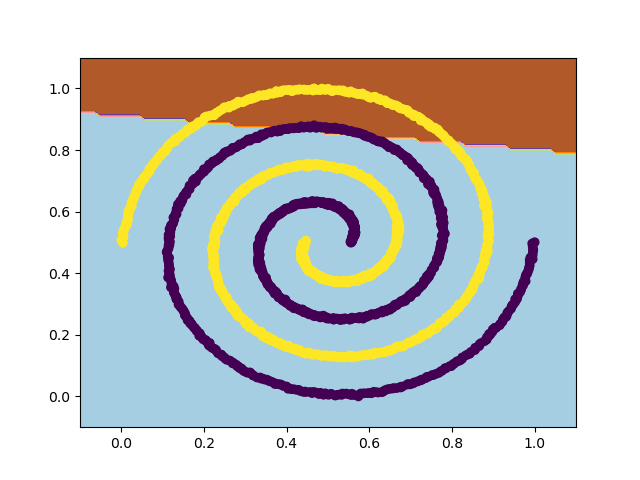}}
    \caption{LC -- Width 2}
    \label{fig:5b}
\end{subfigure}

\begin{subfigure}{0.35\columnwidth}
    \centerline{\includegraphics[width=\columnwidth]{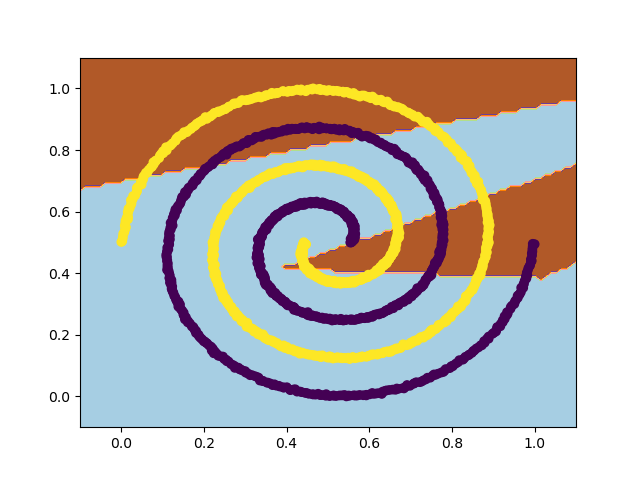}}
    \caption{LC -- Width 10}
    \label{fig:5c}
\end{subfigure}
\hfill%
\begin{subfigure}{0.35\columnwidth}
    \centerline{\includegraphics[width=\columnwidth]{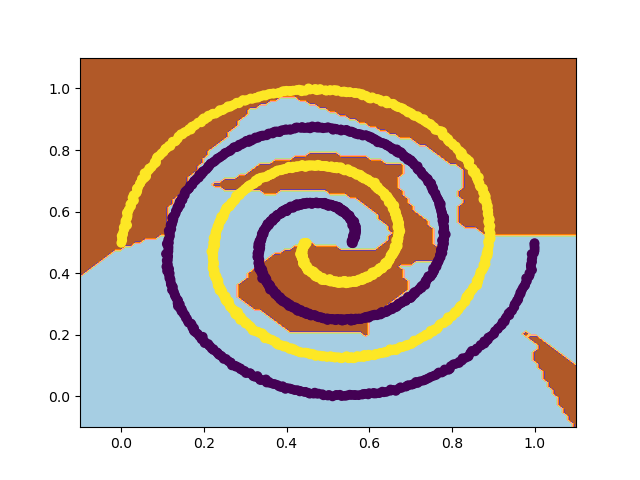}}
    \caption{LC -- Width 100}
    \label{fig:5d}
\end{subfigure}

\begin{subfigure}{0.35\columnwidth}
    \centerline{\includegraphics[width=\columnwidth]{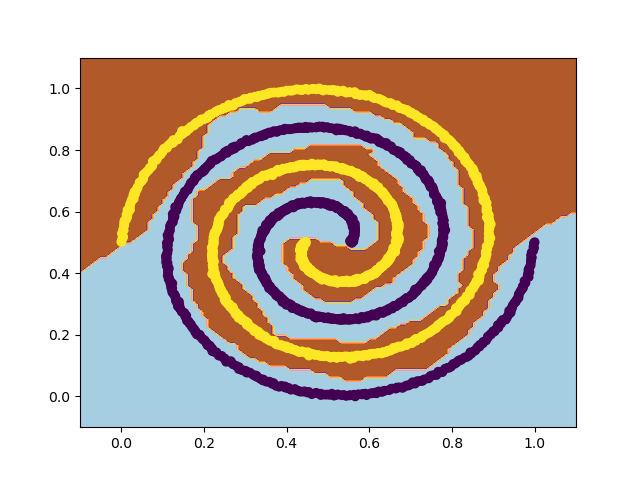}}
    \caption{LC -- Width 500}
    \label{fig:5e}
\end{subfigure}%
\hfill%
\begin{subfigure}{0.35\columnwidth}
    \centerline{\includegraphics[width=\columnwidth]{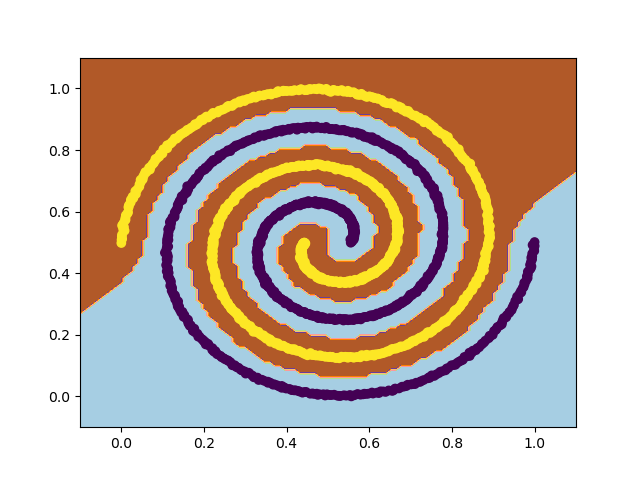}}
    \caption{WS -- Width 2}
    \label{fig:5f}
\end{subfigure}

\begin{subfigure}{0.35\columnwidth}
    \centerline{\includegraphics[width=\columnwidth]{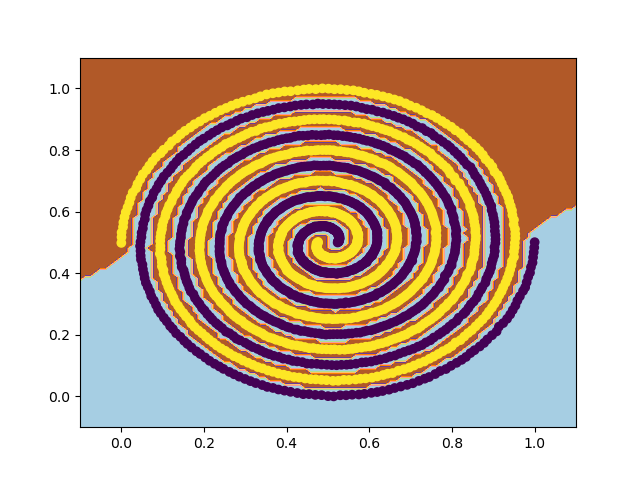}}
    \caption{WS -- Width 2}
    \label{fig:5g}
\end{subfigure}
\hfill%
\begin{subfigure}{0.35\columnwidth}
    \centerline{\includegraphics[width=\columnwidth]{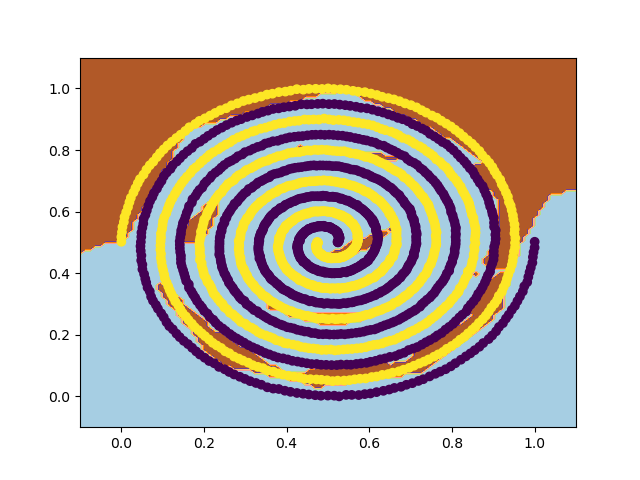}}
    \caption{LC -- Width 5000}
    \label{fig:5h}
\end{subfigure}

\caption{Watershed vs. Linear Classifier as last layer on spiral dataset, illustrated in (a). For a simple spiral dataset, standard linear classifiers need a large number of hidden neurons to estimate the boundary. This is shown in (b)-(e). However, the with watershed classifier, even width $2$ suffices as shown in (f). When the number of revolutions are increased, even 5000 hidden neurons did not suffice for linear classifier as shown in (h). On the other hand, width 2 suffices for watershed classifier, illustrated in (g).}
\label{fig:5}
\end{center}
\vskip -0.2in
\end{figure*}

\crefrange{fig:5b}{fig:5e} shows the boundary for different widths of the hidden layer. Note that one needs at least $100$ hidden nodes to learn the function. When considering $\texttt{N\_REV}=5$, even a width of $5000$ did not suffice (as illustrated in \cref{fig:5h}). However, even width $2$ suffices for the watershed classifier, irrespective of the $\texttt{N\_REV}$. This is illustrated in \cref{fig:5f,fig:5g}.

\section{Experimental Details for \cref{ssec:large_network}}
\label{sec:exp_largenetwork}

\paragraph{Network Architectures}:
\begin{itemize}
    \item[1.] \textsc{MR18}: This is the \textsc{ResNet18} architecture. The implementation is taken from \url{https://github.com/kuangliu/pytorch-cifar/blob/master/models/resnet.py}. 
    \item[2.] \textsc{M3LC}: We consider 3 convolution layers with width $256$ and $64$, with kernel size $5$ and alternating maxpool layers with kernel size $2$ and stride $2$. The linear classifier has an additional classification layer on top of this.
    \item[3.] \textsc{M3LFF}: We consider 3 fully connected feed-forward networks with width $256$ and $64$. The linear classifier has an additional classification layer on top of this. 
\end{itemize}

\paragraph{Training Details:} All experiments are run using A6000 Nvidia GPU with $16$ GB memory. We use \textsc{Adam} optimizer with the learning rate $3e-4$. Experiments with learning rate schedules and other optimizers gave similar results. For watershed results, we run the experiment with $N\_SEEDS=1,5,10,20,40,100$ and report the best performance. We use $N=256$ and $B=2040$ for all the experiments. We use \texttt{early\_stopping} with $\texttt{patience}=20$ epochs on the validation data. We use $80:20$ train/valid split. 

\paragraph{Timing Details:} Using the batch size $B=2040$ with $N=256$ takes around 14 seconds on the system above with \textsc{M3LC} and \textsc{M3LFF} networks, and takes around $70 seconds$ on \textsc{MR18} backbone.

\end{document}